\documentclass[runningheads]{llncs}
\usepackage{graphicx}
\usepackage{comment}
\usepackage{amsmath,amssymb} 
\usepackage{color}
\usepackage{cite}
\usepackage{multirow}
\newcommand{\etal}{\textit{et al.}}
\newcommand{\ie}{\textit{i.e. }}
\newcommand{\eg}{\textit{e.g.}}

\begin{document}
\pagestyle{headings}
\mainmatter
\def\ECCVSubNumber{2400}  

\title{Visual Compositional Learning for Human-Object Interaction Detection} 

%
\author{Zhi Hou\inst{1,2}\and
Xiaojiang Peng\inst{2}\and
Yu Qiao\inst{2} \thanks{corresponding author} \and
Dacheng Tao\inst{1}}
%
%
\institute{
UBTECH Sydney AI Centre, School of Computer Science, Faculty of Engineering, The University of Sydney, Darlington, NSW 2008, Australia\\
\email{zhou9878@uni.sydney.edu.au,dachengtao@sydney.edu.au} \and
Shenzhen Key Lab of Computer Vision and Pattern Recognition, Shenzhen
Institutes of Advanced Technology, Chinese Academy of Sciences \\
\email{\{xj.peng, yu.qiao\}@siat.ac.cn}
}
%

\maketitle

\begin{abstract}

  Human-Object interaction (HOI) detection aims to localize and infer relationships between human and objects in an image. It is challenging because an enormous number of possible combinations of objects and verbs types forms a long-tail distribution. We devise a deep Visual Compositional Learning (VCL) framework, which is a simple yet efficient framework to effectively address this problem. VCL first decomposes an HOI representation into object and verb specific features, and then composes new interaction samples in the feature space via stitching the decomposed features. The integration of decomposition and composition enables VCL to share object and verb features among different HOI samples and images, and to generate new interaction samples and new types of HOI, and thus largely alleviates the long-tail distribution problem and benefits low-shot or zero-shot HOI detection. Extensive experiments demonstrate that the proposed VCL can effectively improve the generalization of HOI detection on HICO-DET and V-COCO and outperforms the recent state-of-the-art methods on HICO-DET. Code is available at \url{https://github.com/zhihou7/VCL}.

\end{abstract}

\section{Introduction}
Human-Object interaction (HOI) detection \cite{chao2018learning, gao2018ican, li2018transferable, qi2018learning} aims to localize and infer relationships (verb-object pairs) between human and objects in images.
The main challenges of HOI come from the complexity of interaction and the long-tail distribution of possible verb-object combinations \cite{chao2018learning, shen2018scaling, xu2019learning}.
In practice, a few types of interactions dominate the HOI samples,
while a large number of interactions are rare which are always difficult to obtain sufficient training samples.

\begin{figure}[t]
\begin{center}
\includegraphics[width=.8\textwidth]{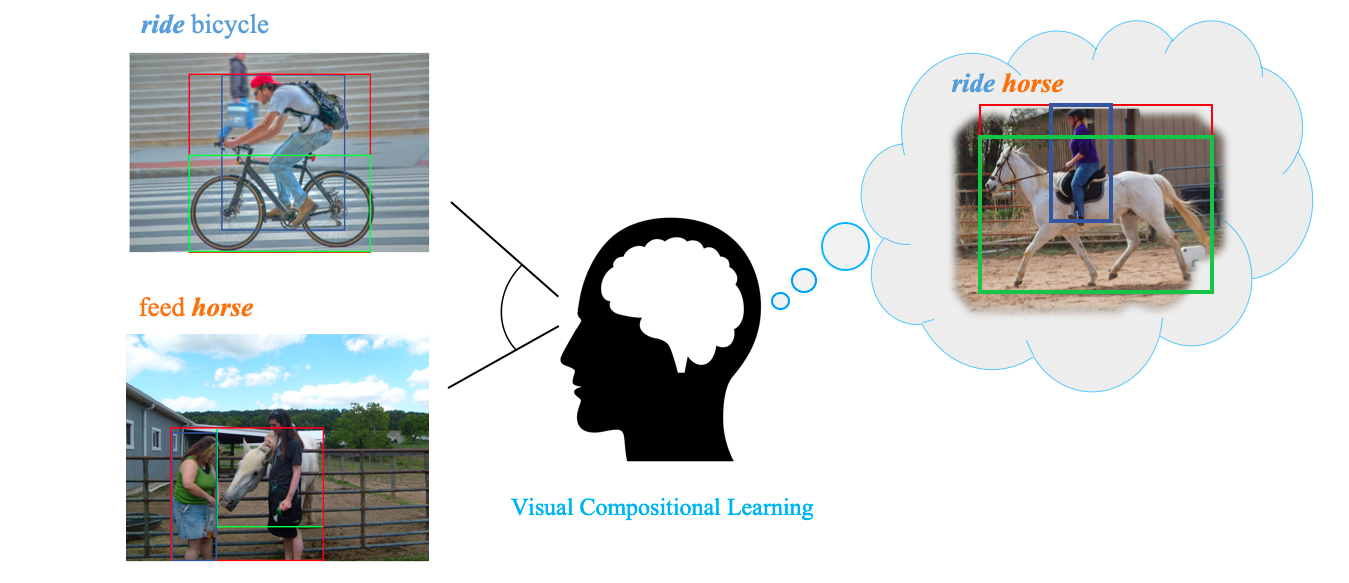}
\end{center}
   \caption{An illustration of Visual Compositional Learning (VCL). VCL constructs the new concept of $\left \langle ride, horse\right \rangle$ from $\left\langle feed, horse\right \rangle$ and $\left \langle ride, bicycle\right\rangle$ via visual compositional learning}
\label{fig:imagine}
\end{figure}

The visual scenes are composed of basic elements, such as objects, parts, and other semantic regions.
It is well-acknowledged that humans perceive world in a compositional way in which visual scenes are treated as a layout of distinct semantic objects \cite{spelke1990principles, hoffman1983parts}. We can understand HOIs by decomposing them into objects and human interaction (verb) types. This decomposition helps to solve the rare Human-Object Interactions with long-tailed distribution. For example, in HICO-DET dataset \cite{chao2018learning}, $\left \langle hug, suitcase \right \rangle$ is a rare case with only one example, while we have more than 1000 HOI samples including object ``suitcase'', and 500 samples including verb ``hug''. Obviously, object representations can be shared among different HOIs. And samples with the same verb usually exhibit similar human pose or action characteristics \cite{xu2019learning}. By compositing the concepts of ``suitcase'' and ``hug'' learned from these large number samples, one can handle the rare case $\left \langle hug, suitcase \right \rangle$. This inspires to reduce the complexity of HOI detection and handle unseen/rare categories via learning compositional components, \ie human verb and objects from visual scenes. Note this idea is near but different from disentangling representation learning \cite{bengio2013representation} (e.g. factors method \cite{shen2018scaling} in HOI detection) which aims to separate the distinct and informative factors from input examples. Similar to disentangling, compositional learning of HOI also includes the decomposing step. Unlike disentangling, compositional learning further composes novel HOI examples with decomposed factors, which is helpful to address low-shot and zero-shot HOI detection.

Inspired by the above analysis, this paper proposes a deep Visual Compositional Learning (VCL) frame work for Human-Object Interaction Detection, which performs compositional learning on visual verb and object representations. VCL simultaneously encourages shared verb and object representation across images and HOI types.
As illustrated in Figure \ref{fig:imagine}, with the semantic features of `\textit{horse}' and `\textit{ride}' in the images of $\left\langle feed, horse\right \rangle$ and $\left \langle ride, bicycle\right\rangle$, one may compose a new interaction feature $\left \langle ride, horse\right \rangle$ using off-the-shelf features.

To perform compositional learning for HOI detection, VCL faces three challenges. Firstly, verb features are usually highly correlated with human and object features. It is non-trivial to decouple the verb representations from those of human and objects in scenes. Unlike prior works~\cite{gao2018ican, gupta2018no}, which extract the verb representations from human boxes, we build verb representation from the union box of human and object. Meanwhile, we share weights between verb and human stream in the mutli-branches structure to emphasize the verb representation on human box region. Our verb representation yields more discriminative cues for the final detection task. Secondly, the number of verbs and objects within a single image is limited for composition. We present a novel feature compositional learning approach by composing HOI samples with verb and object features from different images and different HOI types.
In this way, our VCL encourages the model to learn the shared and distinctive verb and object representations that are insensitive to variations (\ie the specific images and interactions). Thirdly, HOIs always exhibit long-tail distribution where a large number of categories have very few even zero training samples. VCL can compose new interactions and novel types of HOIs in the feature space (\ie verb + object), e.g., the rare HOI $\left \langle wash, cat\right \rangle$ can be drawn from $\left \langle wash, dog\right \rangle$ and $\left \langle feed, cat\right \rangle$.

Overall, our main contributions can be summarized as follows,


\begin{itemize}
\item We creatively devise a deep Visual Compositional Learning framework to compose novel HOI samples from decomposed verbs and objects to relieve low-shot and zero-shot issues in HOI detection. Specifically, we propose to extract verb representations from union box of human and object and compose new HOI samples and new types of HOIs from pairwise images.
\item Our VCL outperforms considerably previous state-of-the-art methods on the largest HOI Interaction detection dataset HICO-DET \cite{chao2018learning}, particularly for rare categories and zero-shot.
\end{itemize}

\section{Related Works}

\subsection{Human-Object Interaction Detection}
Human-Object Interaction \cite{chao2015hico, chao2018learning} is essential for deeper scene understanding. Different from {\bf Visual Relationship Detection} \cite{lu2016visual}, {\bf HOI} is a kind of human-centric relation detection. Several large-scale datasets (V-COCO \cite{gupta2015visual} and HICO-DET \cite{chao2018learning}) were released for the exploration of HOI detection. Chao \etal \cite{chao2018learning} introduced a multi-stream model combining visual features and spatial location features to help tackle this problem. Following the multi-stream structure in \cite{chao2018learning}, Gao \etal \cite{gao2018ican} further exploited an instance centric attention module and Li \etal \cite{li2018transferable} utilized interactiveness to explicitly discriminate non-interactive pairs to improve HOI detection. Recently, Wang \etal \cite{wang2019deep} proposed a contextual attention framework for Human-Object Interaction detection. GPNN \cite{qi2018learning} and RPNN \cite{Zhou_2019_ICCV} were introduced to model the relationships with graph neural network among parts or/and objects. Pose-aware Multi-level Feature Network \cite{wan2019pose} aimed to generate robust predictions on fine-grained human object interaction. Different from the previous works \cite{chao2018learning, gao2018ican, li2018transferable, qi2018learning, wang2019deep, Zhou_2019_ICCV, wan2019pose} who mainly focus on learning better HOI features, we address the long-tailed and zero-shot issues in HOI detection via Visual Compositional Learning.

\subsection{Low-shot and Zero-shot Learning}

Our work also ties with low-shot learning \cite{wang2018low} within long-tailed distribution \cite{liu2019large} and zero-shot learning recognition \cite{xian2017zero}.
Shen \etal \cite{shen2018scaling} introduced a factorized model for HOI detection that disentangles reasoning on verbs and objects to tackle the challenge of scaling HOI recognition to the long tail of categories, which is similar to our work. But we design a compositional learning approach to compose HOI examples. Visual-language joint embedding models \cite{xu2019learning, Peyre_2019_ICCV} enforce the representation of the same verb to be similar among different HOIs by the intrinsic semantic regularities. \cite{Peyre_2019_ICCV} further transfered HOIs embeddings from seen HOIs to unseen HOIs using analogies for zero-shot HOI detection. However, different from \cite{Peyre_2019_ICCV} who aims to zero-shot learning, VCL targets at Generalized Zero-Shot Learning \cite{xian2018zero}. In \cite{bansal2019detecting}, a generic object detector was incorporated to generalize to interactions involving previously unseen objects. Also, Yang \etal \cite{yang2018shuffle} proposed to alleviate the predicate bias to objects for zero-shot visual relationship detection. Similar to previous approaches \cite{shen2018scaling, xu2019learning, bansal2019detecting, Peyre_2019_ICCV}, we also equally treat the same verb from different HOIs. However, all those works \cite{shen2018scaling, xu2019learning, bansal2019detecting, Peyre_2019_ICCV} largely ignore the composition of verbs and objects. In contrast, we propose the Visual Compositional Learning framework to relieve the long-tail and zero-shot issues of HOI detection jointly,
and we demonstrate the efficiency by massive experiments, especially for rare/unseen data in HOI detection.

\subsection{Feature Disentangling and Composing}

Disentangled representation learning has attracted increasing attention in various kinds of visual task \cite{bengio2013representation, higgins2017beta,
locatello2018challenging, higgins2017scan, burgess2019monet}
and the importance of Compositional Learning to build intelligent machines is acknowledged \cite{bengio2013representation, lake2017building, garnelo2019reconciling, battaglia2018relational}.
Higgins \etal \cite{higgins2017scan} proposed Symbol-Concept Association Network (SCAN) to learn  hierarchical visual concepts.
Recently, \cite{burgess2019monet} proposed Multi-Object network (MONet) to decompose scenes by training a Variational Autoencoder together with a recurrent attention network. However, both SCAN \cite{higgins2017scan} and MONet \cite{burgess2019monet} only validate their methods on the virtual datasets or simple scenes.

Besides, Compositional GAN \cite{azadi2018compositional} was introduced to generate new images from a pair of objects.
Recently, Label-Set Operations network (LaSO) \cite{alfassy2019laso} combined features of image pairs to synthesize feature vectors of new label sets according to certain set operations on the label sets of image pairs for multi-label few-shot learning.
Both Compositional GAN \cite{azadi2018compositional} and LaSO \cite{alfassy2019laso}, however, compose the features from two whole images
and depend on generative network or reconstruct loss.
In addition, Kato \etal \cite{kato2018compositional} introduced a compositional learning method for HOI classification \cite{chao2015hico} that utilizes the visual-language joint embedding model to the feature of the whole of image. But \cite{kato2018compositional} did not involve multiple objects detection in the scene.
Our visual compositional learning framework differs from them in following aspects: i) it composes interaction features from regions of images, ii) it simultaneously encourages \textit{discriminative} and \textit{shared} verb and object representations.

\begin{figure*}
\begin{center}
\includegraphics[width=0.8\textwidth, height=5.0cm]{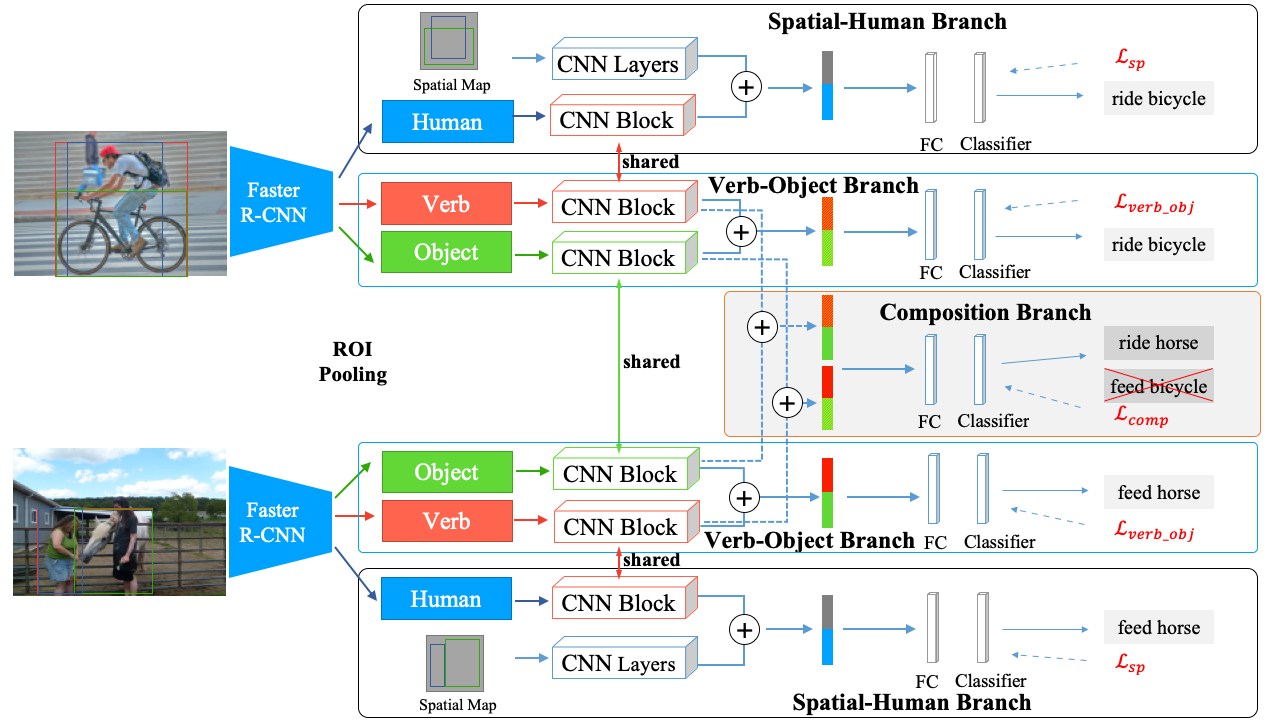}

\end{center}
   \caption{Overview of the proposed Visual Compositional Learning (VCL) framework. Given two images, we first detect human and objects with Faster-RCNN \cite{ren2015faster}. Next, with ROI-Pooling and Residual CNN blocks, we extract human features, verb features (\ie the union box of human and object), and object features. Then, these features are fed into the following branches: individual spatial-human branch, verb-object branch and composited branch. Finally, HOI representations from verb-object branch and composited branch are classified by a shared FC-Classifier, while HOI representations from spatial-human branch are classified by an individual FC-Classifier. \textit{Note that all the parameters are shared across images and the newly composited HOI instances can be from a single image if the image includes multiple HOIs}}
\label{fig:structure}
\end{figure*}

\section{Visual Compositional Learning}
In this section, we present our Visual Compositional Learning (VCL) framework for HOI detection. We first provide an overview of VCL and then detail the HOI branches of VCL. Last, we describe how we compose new HOIs and apply VCL to zero-shot detection.

\subsection{Overview}
To address the long-tail and zero-shot issues of HOI detection, we propose the Visual Compositional Learning (VCL) framework to learn shared object and verb features and compose novel HOI samples with these features. As shown in Figure \ref{fig:structure}, to perform compositional learning, our VCL takes as input a randomly selected image pair. Then we employ a Faster R-CNN \cite{ren2015faster} with backbone ResNet-50 \cite{he2016deep} to detect human and objects in images. Subsequently, we use ROI-Pooling and Residual CNN blocks to obtain features of human, verbs, and objects individually.
Then, to obtain HOI detection, these features together with a human-object spatial map are fed into a spatial-human branch and a verb-object branch. Particularly, composited features are fed into the composited branch for compositional learning.
It is worth noting that all the parameters are shared across images.

\subsection{Multi-branch Network}
\label{sec:multibranch}
Multi-branch architectures are usually used in previous HOI detection works \cite{chao2018learning, gao2018ican,li2018transferable} where each branch processes a kind of input information.
Similarly, VCL includes a spatial-human branch and a verb-object branch. But unlike previous works, VCL has a composition branch which helps the training of HOI detection.

\textbf{Spatial-Human branch.} The spatial-human branch processes human feature and the spatial map of human and object. Following \cite{gao2018ican}, the input of spatial feature is a two-channel 64x64 tensor consisting of a person map and an object map. For the person map, the value of a position will be 1 if it is in in the person box otherwise 0. The object map is similar. We concatenate the spatial map with the human feature.

\textbf{Verb-Object branch.} The verb-object branch in each image includes a verb stream and an object stream. Unlike prior works~\cite{gao2018ican, gupta2018no} which view the human features as verb representations, our newly introduced verb branch extracts \textit{a verb representation from the union box of a human box and an object box}. Meanwhile, similar to \cite{gao2018ican, gupta2018no}, we share the weights between human stream in Spatial-Human branch and verb stream from the union box in Verb-Object branch. Our verb representation is more discriminative which contains more useful contextual information within the union box of human and object.

\textbf{Composition branch.} We compose new verb-object interaction samples in the feature space from the verbs and objects between and within images and then \textit{these synthesized samples are trained jointly with the interactions annotated in the dataset}. It in turn improves the generalization of HOI detection, particularly for rare and unseen HOIs.

\begin{figure*}[t]
\begin{center}
\includegraphics[width=0.9\textwidth]{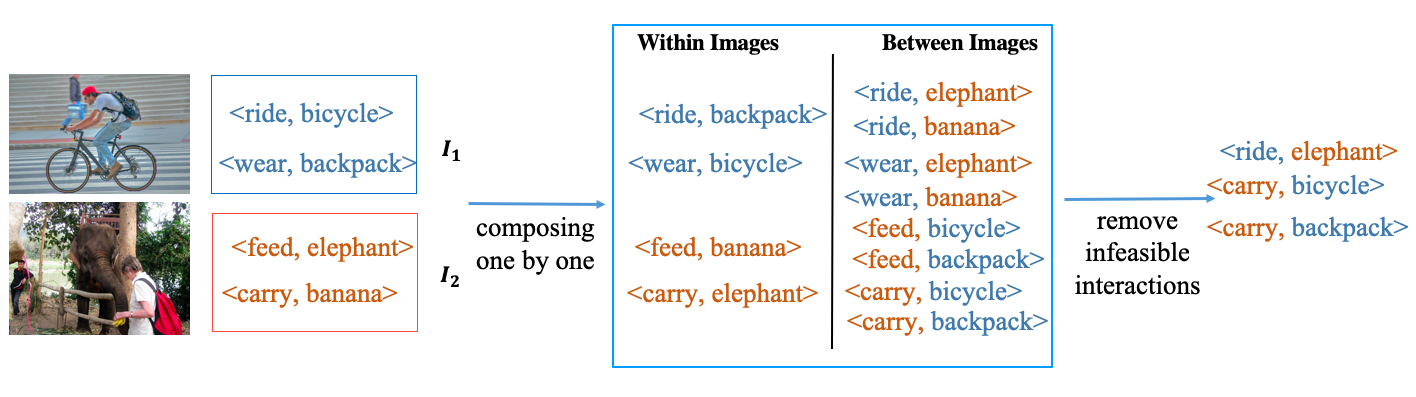}
\end{center}
   \caption{Illustration of the process of composing new interactions. Given two images $I_1$ and $I_2$, we compose new interaction samples within single image and between them by first considering all possible verb-object pairs and then removing infeasible interactions}
\label{fig:composing_type}
\end{figure*}

\subsection{Composing Interactions}
The key idea of our proposed VCL framework is to compose new interaction samples within and between images. This composition process encourages the network to learn shared object and verb features across HOIs by composing massive diverse HOI samples. As shown in Figure \ref{fig:composing_type}, the composition process mainly contains two stages: generating all possible interactions (\ie verb-object pairs) and removing infeasible interactions.
Given two images $I_1$ and $I_2$, we compose new interaction samples within single image and between images by first considering all possible verb-object pairs and then removing infeasible interactions in the HOI label space.

Existing HOI labels mainly contain one object and at least one verb, which set the HOI detection as a multi-label problem. To avoid frequently checking verb-object pairs, we design an efficient composing and removing strategy. First, we decouple the HOI label space into a verb-HOI  matrix $\mathbf{A}_v\in R^{N_v\times C}$ and an object-HOI matrix $\mathbf{A}_o\in R^{N_o\times C}$, where $N_v$, $N_o$, and $C$ denote the number of verbs, objects and HOI categories respectively. $\mathbf{A}_v$ ($\mathbf{A}_o$) can be viewed as the cooccurence matrix between verbs (objects) and HOIs. Then, given binary HOI label vectors $\mathbf{y} \in R^{N \times C}$, where $N$, $C$ denote the number of interactions and HOI categories respectively. we can obtain the object label vector and verb label vector as follows,

\begin{equation}
\mathbf{l}_o = \mathbf{y} \mathbf{A}_o ^\mathsf{T} , ~ \mathbf{l}_v = \mathbf{y} \mathbf{A}_v ^\mathsf{T} ,
\end{equation}
where $\mathbf{l}_o \in R^{N \times N_o}$ is usually one-hot vectors meaning one object of a HOI example, and $\mathbf{l}_v \in R^{N \times N_v}$ is possiblely multi-hot vectors meaning multiple verbs. \eg $\left \langle \{hold, sip\}, cup \right \rangle$). Similarly, we can generate new interactions from arbitrary $\mathbf{l}_o$ and $\mathbf{l}_v$ as follows,

\begin{equation}
\label{formul:label}
\hat{\mathbf{y}} = (\mathbf{l}_o \mathbf{A}_o) \& (\mathbf{l}_v\mathbf{A}_v),
\end{equation}

where $\&$ denotes the ``and'' logical operation. The infeasible HOI labels that do not exist in the given label space are all-zero vectors after the logical operation. And then, we can filter out those inefasible HOIs. In implementation, we obtain verbs and objects from two images by ROI pooling and treat them within and between images as same. Therefore, we do not treat two levels of composition differently during composing HOIs.

\textbf{Zero-Shot Detection}. The composition process makes VCL handling zero-shot detection naturally. Specifically, with the above-mentioned operation, we can generate HOI samples for zero-shot (in the given HOI label space) between and within images which may not be annotated in the training set.

\subsection{Training and Inference}
\textbf{Training}. We train the proposed VCL in an end-to-end manner with Cross Entropy (CE) losses from multiple branches: $\mathcal{L}_{sp}$ from the spatial-human branch for original HOI instances, $\mathcal{L}_{verb\_obj}$ from verb-object branch for original HOI instances, and $\mathcal{L}_{comp}$ from verb-object branch for composited HOI instances. Formally, the total training loss is defined as follows,
\begin{align}
  \mathcal{L} = \mathcal{L}_{sp} + \lambda_1 \mathcal{L}_{verb\_obj} + \lambda_2 \mathcal{L}_{comp},
\end{align}
where $\lambda_1$ and $\lambda_2$ are two hyper-parameters. We employ the composition process (\ie composing new HOI instances) in each minibatch at training stage.

\textbf{Inference}. At test stage, we remove the composition operation and use the spatial-human branch and the verb-object branch to recognize interaction (\ie human-object pair) of an input image. We predict HOI scores in a similar manner to \cite{gao2018ican}. For each human-object bounding box pair $(b_h, b_o)$, we predict the score $S^c_{h,o}$ for each category $c \in {1, ..., C}$, where $C$ denotes the total number of possible HOI categories. The score $S^c_{h,o}$ depends on the confidence for the individual object detection scores ($s_h$ and $s_o$) and the interaction prediction based on verb-object branch $s^c_{verb\_obj}$ and spatial-human branch $s^c_{sp}$. Specifically, Our final HOI score $S^c_{h,o}$ for the human-object bounding box pair $(b_h, b_o)$ is :
\begin{equation}
\label{eq:fuse_scores}
S^c_{h,o} = s_h \cdot s_o \cdot s^c_{verb\_obj} \cdot s^c_{sp}
\end{equation}

\section{Experiment}
In this section, we first introduce datasets and metrics and then provide the details of the implementation of our method. Next, we report our experimental results compared with state-of-the-art approaches and zero-shot results. Finally, we conduct ablation studies to validate the components in our framework.

\subsection{Datasets and Metrics}
{\bf Datasets}. we adopt two HOI datasets HICO-DET \cite{chao2018learning} and V-COCO \cite{gupta2015visual}.
 HICO-DET \cite{chao2018learning} contains 47,776 images (38,118 in train set and 9,658 in test set),
600 HOI categories constructed by 80 object categories and 117 verb classes. HICO-DET provides more than 150k
 annotated human-object pairs. V-COCO \cite{gupta2015visual} provides 10,346 images (2,533 for
training, 2,867 for validating and 4,946 for testing) and 16,199 person instances.
Each person has annotations for 29 action categories and there are no interaction labels including objects.

{\bf Metrics.} We follow the settings in \cite{gao2018ican}, \ie a prediction is a true positive
only when the human and object bounding boxes both have IoUs larger than 0.5 with reference to ground
truth and the HOI classification result is accurate. We use the role mean average precision to
measure the performance on V-COCO.

\subsection{Implementation Details}

For HICO-DET dataset, we utilize the provided interaction labels to decompose and compose interactions labels.
For V-COCO dataset, which only provides verb (\ie action) annotations for COCO images \cite{lin2014microsoft},
we obtain object labels (only 69 classes in V-COCO images) from the COCO annotations.
Therefore, we have 69 classes of objects and 29 classes of actions in V-COCO that construct 238 classes of
Human-Object pairs to facilitate the detection of 29 classes of actions with VCL.
In addition, following the released code of \cite{li2018transferable},
we also apply the same reweighting strategy in HICO-DET and V-COCO,
and we keep it for the composited interactions. Besides, to prevent composited interactions from dominating the training of the model,
we randomly select composited interactions in each minibatch to maintain the same number of composited interactions as we do of non-composited interactions.

For a fair comparison, we adopt the object detection results
and pre-trained weights provided by authors of \cite{gao2018ican}. We apply two 1024-d fully-connected layers to classify the
interaction feature concatenated by verb and object. We train our network for 1000k iterations on the
HICO-DET dataset and 500k iterations on V-COCO dataset with an initial learning rate of 0.01, a weight decay of 0.0005, and a momentum of 0.9.
We set $\lambda_1$ of 2 and $\lambda_2$ of 0.5 for HICO-DET dataset,
while for V-COCO dataset, $\lambda_1$ is 0.5 and $\lambda_2$ is 0.1. In order to compose enough new interactions between images for training,
we increase the number of interactions in each minibatch while reducing the number of augmentations for each interaction to keep the batch size unchanged.
All experiments are conducted on a single Nvidia GeForce RTX 2080Ti GPU with Tensorflow.

\subsection{Results and Comparisons}

We compare our proposed Visual Compositional Learning with
those state-of-the-art HOI detection approaches \cite{qi2018learning, gao2018ican, li2018transferable, gupta2018no, wang2019deep, xu2019learning, wan2019pose, Peyre_2019_ICCV, bansal2019detecting}
on HICO-DET. The HOI detection result is evaluated with mean average precision (mAP) (\%). We use the settings: Full (600 HOIs), Rare (138 HOIs), Non-Rare (462 HOIs) in Default mode and Known Object mode on HICO-DET.

{\bf Comparisons with state-of-the-art.} From Table~\ref{table:sota_hico}, we can find that we respectively improve the performance of Default and Know modes with Full setting to 19.43\% and 22.00\% without external knowledge.
In comparison with \cite{wan2019pose} who achieves the best mAP in Rare category on HICO-DET among the previous works,
we improve the mAP by 1.97\% in Full category and by 0.9\% in Rare category. Particularly, we do not use pose information like \cite{wan2019pose}. We improve dramatically by over {\bf 3\%} in Rare category compared to other visual methods \cite{qi2018learning, gao2018ican, li2018transferable, gupta2018no, Zhou_2019_ICCV}. Besides, we achieve 1.92\% better result than \cite{Peyre_2019_ICCV} in the rare category although we only use the visual information.

Particularly, \cite{bansal2019detecting} incorporates ResNet101 as their backbone and finetune the object detector on HICO-DET. When we the same backbone and finetune the object detector on HICO-DET, we can largely improve the performance to {\bf 23.63\%}. This also illustrate the great effect of object detector on current two-stage HOI detection method.
\textit{in the two-stage HOI detection, we should not only focus on HOI recognition such as \cite{gao2018ican, li2018transferable, Zhou_2019_ICCV}, but also improve the object detector performance}.

\setlength{\tabcolsep}{4pt}
\begin{table}
\begin{center}
\caption{Comparisons with the state-of-the-art approaches on HICO-DET dataset \cite{chao2018learning}. Xu \etal \cite{xu2019learning}, Peyre \etal \cite{Peyre_2019_ICCV} and Bansal \etal \cite{bansal2019detecting} utilize language knowledge. We include the results of \cite{bansal2019detecting} with the same COCO detector as ours. * means we use the res101 backbone and finetune the object detector on HICO-DET dataset like \cite{bansal2019detecting}
}
\label{table:sota_hico}

\begin{tabular}{@{}lcccccc@{}}
\hline
\multirow{2}{*}{Method}&\multicolumn{3}{c}{Default}&\multicolumn{3}{c}{Known Object}\cr
& Full & Rare & NonRare & Full & Rare & NonRare\\

\hline\hline
GPNN\cite{qi2018learning}  & 13.11 & 9.34 & 14.23 & - & - & -  \\
iCAN\cite{gao2018ican}  &14.84 & 10.45 & 16.15 & 16.26 & 11.33 & 17.73  \\
Li \etal\cite{li2018transferable}  & 17.03 & 13.42 & 18.11 & 19.17 & 15.51 & 20.26 \\
Wang \etal \cite{wang2019deep}  & 16.24 & 11.16 & 17.75 & 17.73 & 12.78 & 19.21 \\
Gupta \etal \cite{gupta2018no}  & 17.18 & 12.17 & 18.68 & - & - & -  \\
Zhou \etal \cite{Zhou_2019_ICCV}  & 17.35 & 12.78 & 18.71 & - & - & -  \\
PMFNet \cite{wan2019pose}  & 17.46 & 15.65 & 18.00 & 20.34 & 17.47 & 21.20 \\
\hline
Xu \etal \cite{xu2019learning} & 14.70 & 13.26 & 15.13 & - & - & - \\
Peyre \etal \cite{Peyre_2019_ICCV} & 19.40 & 14.63 & {\bf 20.87} & - & - & -  \\
Bansal \etal\cite{bansal2019detecting} & 16.96 & 11.73 & 18.52 & - & - & - \\
Bansal* \etal\cite{bansal2019detecting} & 21.96 & 16.43 & 23.62 & - & - & - \\
\hline

Ours (VCL) & {\bf 19.43} & {\bf 16.55} & 20.29 & {\bf 22.00} & {\bf 19.09} & {\bf 22.87} \\
Ours* (VCL) & {\bf 23.63} & {\bf 17.21} & {\bf 25.55} & {\bf 25.98} & {\bf 19.12} & {\bf 28.03} \\

\hline
\end{tabular}
\end{center}
\end{table}
\setlength{\tabcolsep}{1.4pt}



\begin{figure}
 \begin{center}
 \includegraphics[height=3.3cm]{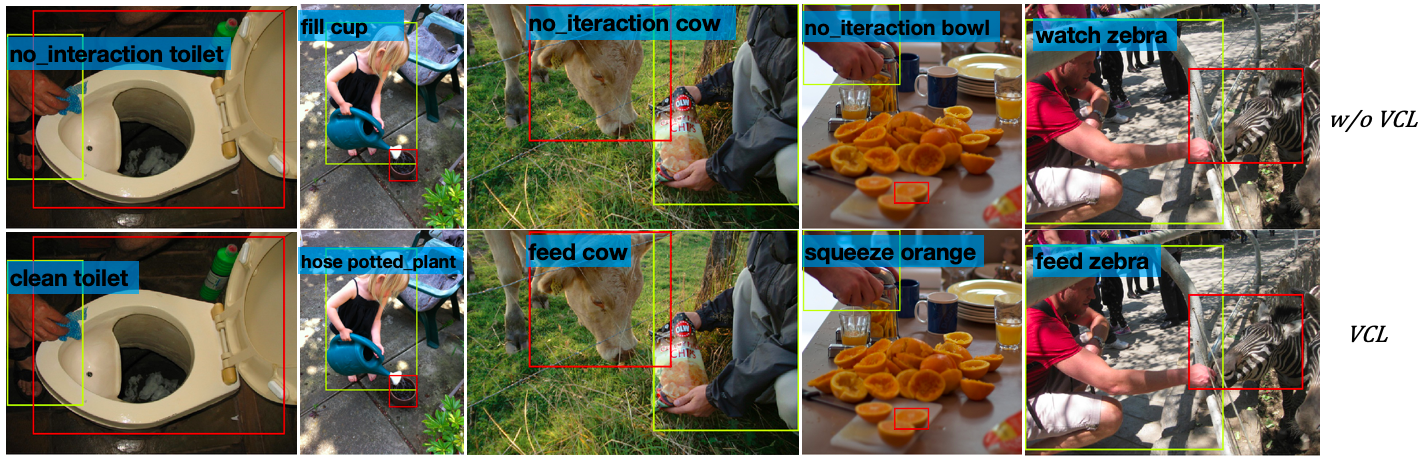}
 \end{center}
    \caption{Some rare HOI detections (Top 1 result) detected by the proposed Compositional Learning and the model without Compositional Learning. The first row is the results of baseline model without VCL. The second row is the results of VCL}
 \label{fig:visual_rare}
 \end{figure}

{\bf Visualization.} In Figure~\ref{fig:visual_rare}, we qualitatively show that our proposed Visual Compositional Learning framework can detect those rare interactions correctly
while the baseline model without VCL misclassifies on HICO-DET.

\subsection{Generalized Zero-shot HOI Detection}
We also evaluate our method in HOI zero-shot detection since our method can naturally be applied to zero-shot detection.
Following \cite{shen2018scaling}, we select 120 HOIs from HOI label set and make sure that the remaining contains all objects and verbs.
The annotations on HICO-DET, however, are long-tail distributed.
The result of selecting preferentially rare labels is different from that of selecting non-rare labels
and we do not know the specific partition of zero shot in \cite{shen2018scaling, bansal2019detecting}.
For a fair comparison, we split the HOI zero-shot experiment into two groups: rare first selection and non-rare first selection.
Rare first selection means we pick out as many rare labels as possible for unseen classes according to the number of instances for zero-shot (remain 92,705 training instances), while non-rare first selection is that we select as many non-rare labels as possible for unseen classes (remain 25,729 training instances). The unseen HOI detection result is evaluated with mean average precision (mAP) (\%). We report our result in the settings: Unseen (120 HOIs), Seen (480 HOIs), Full (600 HOIs) in Default mode and Known Object mode on HICO-DET.

\setlength{\tabcolsep}{4pt}
\begin{table}
\begin{center}
\caption{Comparison of Zero Shot Detection results of our proposed Visual Compositional Learning framework. * means we uses the res101 backbone and finetune object detector on HICO-DET. Full means all categories including Seen and Unseen
  }
\label{table:zero_shot1}
\small
\begin{tabular}{@{}lcccccc@{}}
\hline

\multirow{2}{*}{Method} &
\multicolumn{3}{c}{Default}&\multicolumn{3}{c}{Known Object}\cr\cline{2-7}
&Unseen&Seen&Full&Unseen&Seen&Full \cr

\hline\hline
Shen \etal \cite{shen2018scaling} & 5.62 & - & 6.26 & - & - & -\\
Bansal* \etal \cite{bansal2019detecting} & 10.93 & 12.60 & 12.26 & - & - & - \\
\hline
w/o VCL (rare first)  & 3.30 & 18.63 & 15.56 & 5.53 & 21.30 & 18.15\\
w/o VCL (non-rare first) & 5.06 & 12.77 & 11.23 & 8.81 & 15.37 & 14.06\\
\hline
VCL (rare first) & {\bf 7.55} & {\bf 18.84}  & {\bf 16.58} & {\bf 10.66} & {\bf 21.56} & {\bf 19.38}\\
VCL (non-rare first) & {\bf 9.13} & {\bf 13.67}  & {\bf 12.76} & {\bf 12.97} & {\bf 16.31} & {\bf 15.64} \\
\hline
VCL* (rare first) & {\bf 10.06} & {\bf 24.28}  & {\bf 21.43} & {\bf 12.12} & {\bf 26.71} & {\bf 23.79}\\
VCL* (non-rare first) & {\bf 16.22} & {\bf 18.52}  & {\bf 18.06} & {\bf 20.93} & {\bf 21.02} & {\bf 20.90} \\
\hline
\end{tabular}
\end{center}
\end{table}
\setlength{\tabcolsep}{1.4pt}

We can find in Table~\ref{table:zero_shot1}, both selection strategies witness a consistent increase in all categories compared to the corresponding baseline. Noticeably,
both kinds of selecting strategies witness a surge of performance by {\bf over 4\%} than baseline in unseen category with VCL.
Particularly, our baseline without VCL still achieves low results (3.30\% and 5.06\%) because we predict 600 classes (including 120 unseen categories) in baseline. The model could learn the verb and object of unseen HOIs individually from seen HOIs. Meanwhile, non-rare first selection has more verbs and objects of unseen HOIs individually from seen HOIs than that of rare first selection. Therefore, non-rare first selection has better baseline than rare first selection.
Besides, we improve \cite{shen2018scaling, bansal2019detecting} largely nearly among all categories. In detail, for a fair comparison to \cite{bansal2019detecting}, we also use Resnet-101 and finetune the object detector on HICO-DET dataset, which can largely improve the performance.
This demonstrates that our VCL effectively improves the detection of unseen interactions and maintains excellent HOI detection performance for seen interactions at the same time.

\subsection{Ablation Analysis}

To evaluate the design of our VCL, we first conduct ablation studies about VCL and verb representation on HICO-DET dataset and V-COCO dataset. For V-COCO, we evaluate $AP_{role}$ (24 actions with roles) following \cite{chao2018learning, gupta2015visual}. Besides, we evaluate two branches and composing interactions between images and/or within images on HICO-DET. See supplementary material for more analysis.

\setlength{\tabcolsep}{4pt}
\begin{table}
\begin{center}
  \caption{Ablation study of the proposed Visual Compositional Learning framework on HICO-DET and V-COCO test set.
  VCL means Visual Compositional Learning and Union Verb means we learn verb feature from union box of human and object. Sharing W means sharing weights between human stream and verb stream for verb representation. Re-weighting means we use re-weighting strategy in the code of \cite{li2018transferable}
   }
  \label{table:verbs}
\small
\begin{tabular}{@{}cccccccc@{}}
\hline
\multirow{2}{*}{VCL} & \multirow{2}{*}{Union Verb} & \multirow{2}{*}{Sharing W} &\multirow{2}{*}{Re-weighting}&  \multicolumn{3}{c}{HICO}&\multicolumn{1}{c}{V-COCO}\cr\cline{5-7}
&&&&Full  &Rare &NonRare &$AP_{role}$\cr

\hline\hline
- & - & \checkmark & - & 16.87 & 10.07 & 18.90 & 46.9\\
\checkmark & - & \checkmark & - & 17.35 & 12.10 & 18.91 & 47.2\\
\checkmark & \checkmark & - &  \checkmark & 18.93 & 15.68 & 19.90 & 47.8\\
- & - & \checkmark  &  \checkmark& 18.03 & 13.62 & 19.35 & 47.4 \\
\checkmark & - &  \checkmark  &  \checkmark& 18.57 & 14.83 & 19.69 & 47.7 \\
-  & \checkmark  & \checkmark  &  \checkmark& 18.43 & 14.14 & 19.71  & 47.5 \\
\checkmark & \checkmark & \checkmark  &  \checkmark & {\bf 19.43} & {\bf 16.55} & {\bf 20.29} & {\bf 48.3} \\

\hline
\end{tabular}
\end{center}
\end{table}
\setlength{\tabcolsep}{1.4pt}

{\bf Visual Compositional Learning} is our core method.
With VCL, we can respectively see an obvious increase by 1.00\% in the Full category from Table~\ref{table:verbs} (row 6 vs row 7).
Particularly, the improvement (2.41\%) in Rare category is considerably better than that in Non-Rare category,
which means the proposed VCL is more beneficial for rare data. Noticeably, the improvement of VCL with verb representation learning is better than that without verb representation, which implies that the more discriminative the verb representation is, the better improvement VCL obtains in HOI detection.
We can see a similar trend of VCL on V-COCO in Table~\ref{table:verbs} where the performance decreases by 0.8\% without VCL.
Noticeably, V-COCO aims to recognize the actions (verbs)
rather than the human-object pairs, which means the proposed VCL is also helpful to learn the discriminative action (verbs) representation. Besides, from row 1 and row 2 in Table~\ref{table:verbs}, we can find the proposed VCL is pluggable to re-weighting strategy \cite{japkowicz2002class} (See details in supplementary materials).

{\bf Verb Representation}, which we learn from the union box of human and object, is considerably helpful for the performance. We evaluate its efficiency by comparing human box and union box of human and object.
From row 5 and row 7 in Table~\ref{table:verbs}, we can find that with the proposed verb representation we improve the performance by 0.86\% on HICO-DET and 0.6\% in V-COCO respectively within VCL, which means learning verb representation from the union box of human and object is more discriminative and advantageous for HOI detection. Particularly, we \textit{share the weights of resnet block between human and verb stream} after ROI pooling in the proposed verb representation method. Table~\ref{table:verbs} (row 3 vs row 7) shows sharing weights helps the model improve the performance largely from 18.93\% to 19.43\%. This may be explained by that the union region contains more noise and the model would emphasize the region of the human in the union box by sharing weights to obtain better verb representation.

\setlength{\tabcolsep}{4pt}
\begin{table}
\begin{center}
\caption{The branches ablation study of the model on HICO-DET test set. Verb-object branch only means we train the model without spatial-human branch.
  }
  \label{table:branch}
\small
\begin{tabular}{lccc}
\hline
Method &  Full  & Rare  & NonRare \\
\hline\hline
Two branches & 19.43 & 16.55 & 20.29 \\
Verb-Object branch only during inference & 16.89 & 15.35 & 17.35 \\
Spatial-Human branch only during inference & 16.13 & 12.39 & 17.24 \\
Verb-Object branch only & 15.77 & 13.35 & 16.49 \\
Verb-Object branch only (w/o VCL) & 15.33 & 10.85 & 16.67 \\
\hline
\end{tabular}
\end{center}
\end{table}
\setlength{\tabcolsep}{1.4pt}
{\bf Branches.} There are two branches in our method and we evaluate their contributions in Table~\ref{table:branch}. Noticeably, we apply VCL to Verb-Object branch during training, while we do not apply VCL to Spatial-Human. By keeping one branch each time on HICO-DET dataset during inference, we can find the verb-object branch makes the larger contribution, particularly for rare category ({\bf 3\%}). This efficiently illustrates the advantage of VCL for rare categories. But we can improve the performance dramatically from 16.89\% to 19.43\% with Spatial-Human Branch. Meanwhile, we can find the proposed VCL is orthogonal to spatial-human branch from the last two rows in Table~\ref{table:branch}. Noticeably, \textit{by comparing verb-object branch only during inference and verb-object branch only from training, we can find the spatial-human branch can facilitate the optimization of verb-object branch (improving the mAP from 15.77\% to 16.89\%)}. See supplementary materials for more analysis of two branches in zero-shot detection where the performance of verb-object branch with VCL is over 3.5\% better than spatial-human branch in unseen data.

\setlength{\tabcolsep}{4pt}
\begin{table}
\begin{center}
  \caption{Composing Strategies study of VCL on HICO-DET test set
  }
  \label{table:compose_strategy}
\small
\begin{tabular}{lccc}
\hline
Method &  Full (mAP \%)  & Rare (mAP \%)  & NonRare (mAP \%)  \\
\hline\hline
Baseline (w/o VCL) & 18.43 & 14.14 & 19.71 \\
Within images & 18.48 & 14.46 & 19.69 \\
Between images & 19.06 & 14.33 & 20.47 \\
Between and within images & 19.43 & 16.55 & 20.29 \\
\hline
\end{tabular}
\end{center}
\end{table}
\setlength{\tabcolsep}{1.4pt}

{\bf Composing interactions within and/or between images.} In Table~\ref{table:compose_strategy}, we can find composing interaction samples between images is beneficial for HOI detection, whose performance in the Full category increases to 19.06\% mAP, while composing interaction samples within images has similar results to baseline. It might be because the number of images including multiple interactions is few on HICO-DET dataset.
Remarkably, composing interaction samples within and between images notably improves the performance up to 19.43\% mAP in Full and {\bf 16.55\%} mAP in Rare respectively. Those results mean composing interactions within images and between images is more beneficial for HOI detection.

\subsection{Visualization of features}
\label{sec:dis}

\begin{figure*}
 \begin{center}
 \includegraphics[height=4.4cm]{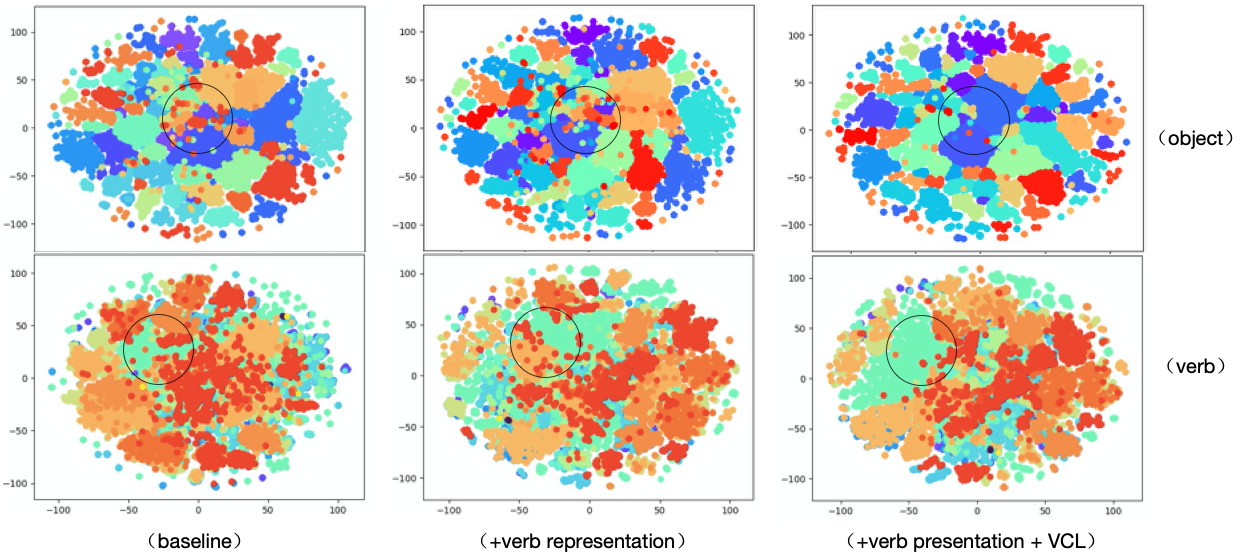}
 \end{center}
    \caption{Visual illustration of object features (80 classes) (up) and verb features (117 classes) (bottom) on HICO-DET dataset (20000 samples) via t-SNE visualization \cite{maaten2008visualizing}. Left is the visual illustration of baseline, middle includes verb representation and the right uses both VCL and verb representation}
 \label{fig:dis}
 \end{figure*}

We also illustrate the verb and object features by t-SNE visualization \cite{maaten2008visualizing}.
Figure~\ref{fig:dis} illustrates that VCL overall improves the the discrimination of verb and object features. There are many noisy points (see black circle region) in Figure~\ref{fig:dis} without VCL and verb presentation. Meanwhile, we can find the proposed verb representation learning is helpful for verb feature learning by comparing verb t-SNE graph between the left and the middle. Besides, the object features are more discriminative than verb. We think it is because the verb feature is more abstract and complex and the verb representation requires further exploration.

\section{Conclusion}

In this paper, we propose a simple yet efficient deep Visual Compositional Learning framework, which composes the interactions from the shared verb and object latent representations between images and within images, to relieve the long-tail distribution issue and zero-shot learning in human-object interaction detection. Meanwhile, we extract a more discriminative verb representation from the union box of human and object rather than human box. Lastly, we evaluate the efficiency of our model on two HOI detection benchmarks, particularly for low-shot and zero-shot detection.

{\bf Acknowledge} This work is partially supported by Science and Technology Service Network Initiative of Chinese Academy of Sciences (KFJ-STS-QYZX-092), Guangdong Special Support Program (2016TX03X276), National Natural Science Foundation of China (U1813218, U1713208), Shenzhen Basic Research Program (JCYJ20170818164704758, CXB201104220032A), the Joint Lab of CAS-HK, Australian Research Council Projects (FL-170100117).

\clearpage
\section*{Appendix}

\section*{A The two branches study in zero-shot HOI detection}
\label{sec:zero_shot_branches}

\begin{table}
\begin{center}
\caption{Two branches ablation study of the proposed Visual Compositional Learning framework in zero-shot HOI detection on HICO-DET test set during inference.
}
\label{table:branch_zero_shot}
\small
\begin{tabular}{lccc}
\hline
Method & Unseen & Seen & Full \\
\hline\hline
Verb-Object branch (rare first) & {\bf 7.85} & 15.48 & 13.95 \\
Spatial-Human branch (rare first) & 4.33 & 15.92 & 13.60 \\
Two branches (rare first) & 7.55 & {\bf 18.84} & {\bf 16.58} \\
\hline
Verb-Object branch (non-rare first) & {\bf 10.61} & 10.95 & 10.88 \\
Spatial-Human branch (non-rare first) & 5.71 & 11.82 & 10.60 \\
Two branches (non-rare first) & 9.13 & {\bf 13.67}  & {\bf 12.76} \\

\hline
\end{tabular}
\end{center}
\end{table}

We evaluate the contribution of the two branches in zero-shot HOI detection. From Table~\ref{table:branch_zero_shot}, we can find the performance of verb-object branch in Seen category and Full category is similar to that of spatial-human branch, while verb-object branch is 3.52\% and {\bf 4.90\%} better than spatial branch in selecting rare first and selecting non-rare first respectively in the Unseen category. {\bf Particularly, after we fuse the result of the two branches, the Unseen category witnesses a considerable decrease in the two selecting strategies.} This illustrates that the additional spatial-human branch contributes to the full performance while \textit{the verb-object branch with VCL efficiently benefits the zero-shot recognition.}

\section*{B The effect of the number of interactions in minibatch}
\label{sec:interaction_nums}

In order to compose enough interactions for Visual Compositional Learning, we increase the number of interactions in each minibatch while reducing the number of augmentations for each interaction and the number of negative interactions. Therefore, we can still optimize the network in a single GPU. From Table~\ref{table:effect_iteraction_nums}, we can find the baseline model of different iteractions has similar results with 18.43 mAP and 18.47 mAP respectively. However, we witness a better improvement (1.0 mAP vs 0.44 mAP) if we increase the interaction classes in the minibatch.

\begin{table}
\begin{center}
\caption{The results of the number of interactions in minibatch in HICO-DET.
}
\label{table:effect_iteraction_nums}
\small
\begin{tabular}{ccccc}
\hline
the number of interactions & VCL & Full & Rare & NonRare \\
\hline\hline

1 & - & 18.41 & 14.17 & 19.68 \\
1 & \checkmark & 18.85 & 14.98 & 20.01 \\
5 & - & 18.43 & 14.14 & 19.71 \\
5 & \checkmark & {\bf 19.43} & {\bf 16.55} & {\bf 20.29} \\

\hline
\end{tabular}
\end{center}
\end{table}

\section*{C Hyper-Parameters and Baseline}
\label{sec:hyper-param}

In our proposed framework, there are two hyper-parameters $\lambda_1$ and $\lambda_2$.
We evaluate the performance when we set different values for the two hyper-parameters in Table~\ref{table:lambda_1_comparisons}.

Like \cite{gao2018ican, li2018transferable}, we first detect the objects in the image and then use the object detection results to infer the HOI categories during test. We use the same score threshold (0.8 for human and 0.3 for object ) same as \cite{li2018transferable} in resnet50 coco detector. We use 0.3 for human and 0.1 for object in resnet101 detector that is finetuned on HICO-DET dataset since the finetuned object detection result is largely better.

\begin{table}
\begin{center}
\caption{The results of different values for $\lambda_1$ when $\lambda_2$ is 0.5 and $\lambda_2$ when $\lambda_1$ is 2.0 in HICO-DET .
}
\label{table:lambda_1_comparisons}
\small
\begin{tabular}{lccccc}
\hline
$\lambda_1$ & 1.0 & 1.5 & 2.0 & 2.5 & 3 \\
\hline\hline
Full & 18.96 & 18.95 & 19.43 & 19.29 & 19.34 \\
\hline\hline
$\lambda_2$  & 0.05 & 0.1 & 0.5 & 1.0 & 1.5 \\
\hline
Full & 19.18 & 19.30 & 19.43 & 19.10 & 18.90 \\
\hline
\end{tabular}
\end{center}
\end{table}

We conduct experiments based on the code of \cite{li2018transferable} who released the code in their final version. We find there are two simple but very useful strategies for improvement: reweighting and postprocess for detection. Reweighting is that they allocate different weights for the cross entropy loss according to the number of classes. Postprocess for detection is that we decrease the detection threshold for those images the the detector can't detect any objects and humans. See Table~\ref{table:strategy_1_comparisons} for comparison. See our code for details.

\setlength{\tabcolsep}{4pt}
\begin{table}
\begin{center}
\caption{ Comparison of strategies for baseline}
\label{table:strategy_1_comparisons}
\small
\begin{tabular}{lccc}
\hline
Strategy & Full (mAP \%)  &Rare (mAP \%) &NonRare (mAP \%)\cr
\hline\hline
w/o reweighting &  16.87 & 10.07 & 18.90 \\
w/o postprocess & 17.14 & 12.92 & 18.40 \\
our baseline & 18.03 & 13.62 & 19.35 \\
\hline
\end{tabular}
\end{center}
\end{table}
\setlength{\tabcolsep}{1.4pt}

%
%
%
%

\section*{D Unseen labels}
\label{sec:unseenlabels}

See our released code for the labels of the two selection strategy.

%
%
\bibliographystyle{splncs04}
\bibliography{egbib}

\begin{thebibliography}{10}
\providecommand{\url}[1]{\texttt{#1}}
\providecommand{\urlprefix}{URL }
\providecommand{\doi}[1]{https://doi.org/#1}

\bibitem{alfassy2019laso}
Alfassy, A., Karlinsky, L., Aides, A., Shtok, J., Harary, S., Feris, R.,
  Giryes, R., Bronstein, A.M.: Laso: Label-set operations networks for
  multi-label few-shot learning. In: Proceedings of the IEEE Conference on
  Computer Vision and Pattern Recognition. pp. 6548--6557 (2019)

\bibitem{azadi2018compositional}
Azadi, S., Pathak, D., Ebrahimi, S., Darrell, T.: Compositional gan: Learning
  conditional image composition. arXiv preprint arXiv:1807.07560  (2018)

\bibitem{bansal2019detecting}
Bansal, A., Rambhatla, S.S., Shrivastava, A., Chellappa, R.: Detecting
  human-object interactions via functional generalization. arXiv preprint
  arXiv:1904.03181  (2019)

\bibitem{battaglia2018relational}
Battaglia, P.W., Hamrick, J.B., Bapst, V., Sanchez-Gonzalez, A., Zambaldi, V.,
  Malinowski, M., Tacchetti, A., Raposo, D., Santoro, A., Faulkner, R., et~al.:
  Relational inductive biases, deep learning, and graph networks. arXiv
  preprint arXiv:1806.01261  (2018)

\bibitem{bengio2013representation}
Bengio, Y., Courville, A., Vincent, P.: Representation learning: A review and
  new perspectives. IEEE transactions on pattern analysis and machine
  intelligence  \textbf{35}(8),  1798--1828 (2013)

\bibitem{burgess2019monet}
Burgess, C.P., Matthey, L., Watters, N., Kabra, R., Higgins, I., Botvinick, M.,
  Lerchner, A.: Monet: Unsupervised scene decomposition and representation.
  arXiv preprint arXiv:1901.11390  (2019)

\bibitem{chao2018learning}
Chao, Y.W., Liu, Y., Liu, X., Zeng, H., Deng, J.: Learning to detect
  human-object interactions. In: 2018 ieee winter conference on applications of
  computer vision (wacv). pp. 381--389. IEEE (2018)

\bibitem{chao2015hico}
Chao, Y.W., Wang, Z., He, Y., Wang, J., Deng, J.: Hico: A benchmark for
  recognizing human-object interactions in images. In: Proceedings of the IEEE
  International Conference on Computer Vision. pp. 1017--1025 (2015)

\bibitem{gao2018ican}
Gao, C., Zou, Y., Huang, J.B.: ican: Instance-centric attention network for
  human-object interaction detection. arXiv preprint arXiv:1808.10437  (2018)

\bibitem{garnelo2019reconciling}
Garnelo, M., Shanahan, M.: Reconciling deep learning with symbolic artificial
  intelligence: representing objects and relations. Current Opinion in
  Behavioral Sciences  \textbf{29},  17--23 (2019)

\bibitem{gupta2015visual}
Gupta, S., Malik, J.: Visual semantic role labeling. arXiv preprint
  arXiv:1505.04474  (2015)

\bibitem{gupta2018no}
Gupta, T., Schwing, A., Hoiem, D.: No-frills human-object interaction
  detection: Factorization, appearance and layout encodings, and training
  techniques. arXiv preprint arXiv:1811.05967  (2018)

\bibitem{he2016deep}
He, K., Zhang, X., Ren, S., Sun, J.: Deep residual learning for image
  recognition. In: Proceedings of the IEEE conference on computer vision and
  pattern recognition. pp. 770--778 (2016)

\bibitem{higgins2017beta}
Higgins, I., Matthey, L., Pal, A., Burgess, C., Glorot, X., Botvinick, M.,
  Mohamed, S., Lerchner, A.: beta-vae: Learning basic visual concepts with a
  constrained variational framework. ICLR  \textbf{2}(5), ~6 (2017)

\bibitem{higgins2017scan}
Higgins, I., Sonnerat, N., Matthey, L., Pal, A., Burgess, C.P., Bosnjak, M.,
  Shanahan, M., Botvinick, M., Hassabis, D., Lerchner, A.: Scan: Learning
  hierarchical compositional visual concepts. arXiv preprint arXiv:1707.03389
  (2017)

\bibitem{hoffman1983parts}
Hoffman, D.D., Richards, W.: Parts of recognition  (1983)

\bibitem{japkowicz2002class}
Japkowicz, N., Stephen, S.: The class imbalance problem: A systematic study.
  Intelligent data analysis  \textbf{6}(5),  429--449 (2002)

\bibitem{kato2018compositional}
Kato, K., Li, Y., Gupta, A.: Compositional learning for human object
  interaction. In: Proceedings of the European Conference on Computer Vision
  (ECCV). pp. 234--251 (2018)

\bibitem{lake2017building}
Lake, B.M., Ullman, T.D., Tenenbaum, J.B., Gershman, S.J.: Building machines
  that learn and think like people. Behavioral and brain sciences  \textbf{40}
  (2017)

\bibitem{li2018transferable}
Li, Y.L., Zhou, S., Huang, X., Xu, L., Ma, Z., Fang, H.S., Wang, Y.F., Lu, C.:
  Transferable interactiveness prior for human-object interaction detection.
  arXiv preprint arXiv:1811.08264  (2018)

\bibitem{lin2014microsoft}
Lin, T.Y., Maire, M., Belongie, S., Hays, J., Perona, P., Ramanan, D.,
  Doll{\'a}r, P., Zitnick, C.L.: Microsoft coco: Common objects in context. In:
  European conference on computer vision. pp. 740--755. Springer (2014)

\bibitem{liu2019large}
Liu, Z., Miao, Z., Zhan, X., Wang, J., Gong, B., Yu, S.X.: Large-scale
  long-tailed recognition in an open world. In: Proceedings of the IEEE
  Conference on Computer Vision and Pattern Recognition. pp. 2537--2546 (2019)

\bibitem{locatello2018challenging}
Locatello, F., Bauer, S., Lucic, M., Gelly, S., Sch{\"o}lkopf, B., Bachem, O.:
  Challenging common assumptions in the unsupervised learning of disentangled
  representations. arXiv preprint arXiv:1811.12359  (2018)

\bibitem{lu2016visual}
Lu, C., Krishna, R., Bernstein, M., Fei-Fei, L.: Visual relationship detection
  with language priors. In: European conference on computer vision. pp.
  852--869. Springer (2016)

\bibitem{maaten2008visualizing}
Maaten, L.v.d., Hinton, G.: Visualizing data using t-sne. Journal of machine
  learning research  \textbf{9}(Nov),  2579--2605 (2008)

\bibitem{Peyre_2019_ICCV}
Peyre, J., Laptev, I., Schmid, C., Sivic, J.: Detecting unseen visual relations
  using analogies. In: The IEEE International Conference on Computer Vision
  (ICCV) (October 2019)

\bibitem{qi2018learning}
Qi, S., Wang, W., Jia, B., Shen, J., Zhu, S.C.: Learning human-object
  interactions by graph parsing neural networks. In: Proceedings of the
  European Conference on Computer Vision (ECCV). pp. 401--417 (2018)

\bibitem{ren2015faster}
Ren, S., He, K., Girshick, R., Sun, J.: Faster r-cnn: Towards real-time object
  detection with region proposal networks. In: Advances in neural information
  processing systems. pp. 91--99 (2015)

\bibitem{shen2018scaling}
Shen, L., Yeung, S., Hoffman, J., Mori, G., Fei-Fei, L.: Scaling human-object
  interaction recognition through zero-shot learning. In: 2018 IEEE Winter
  Conference on Applications of Computer Vision (WACV). pp. 1568--1576. IEEE
  (2018)

\bibitem{spelke1990principles}
Spelke, E.S.: Principles of object perception. Cognitive science
  \textbf{14}(1),  29--56 (1990)

\bibitem{wan2019pose}
Wan, B., Zhou, D., Liu, Y., Li, R., He, X.: Pose-aware multi-level feature
  network for human object interaction detection. In: Proceedings of the IEEE
  International Conference on Computer Vision. pp. 9469--9478 (2019)

\bibitem{wang2019deep}
Wang, T., Anwer, R.M., Khan, M.H., Khan, F.S., Pang, Y., Shao, L., Laaksonen,
  J.: Deep contextual attention for human-object interaction detection. arXiv
  preprint arXiv:1910.07721  (2019)

\bibitem{wang2018low}
Wang, Y.X., Girshick, R., Hebert, M., Hariharan, B.: Low-shot learning from
  imaginary data. In: Proceedings of the IEEE conference on computer vision and
  pattern recognition. pp. 7278--7286 (2018)

\bibitem{xian2018zero}
Xian, Y., Lampert, C.H., Schiele, B., Akata, Z.: Zero-shot learning—a
  comprehensive evaluation of the good, the bad and the ugly. IEEE transactions
  on pattern analysis and machine intelligence  \textbf{41}(9),  2251--2265
  (2018)

\bibitem{xian2017zero}
Xian, Y., Schiele, B., Akata, Z.: Zero-shot learning-the good, the bad and the
  ugly. In: Proceedings of the IEEE Conference on Computer Vision and Pattern
  Recognition. pp. 4582--4591 (2017)

\bibitem{xu2019learning}
Xu, B., Wong, Y., Li, J., Zhao, Q., Kankanhalli, M.S.: Learning to detect
  human-object interactions with knowledge. In: Proceedings of the IEEE
  Conference on Computer Vision and Pattern Recognition (2019)

\bibitem{yang2018shuffle}
Yang, X., Zhang, H., Cai, J.: Shuffle-then-assemble: Learning object-agnostic
  visual relationship features. In: Proceedings of the European conference on
  computer vision (ECCV). pp. 36--52 (2018)

\bibitem{Zhou_2019_ICCV}
Zhou, P., Chi, M.: Relation parsing neural network for human-object interaction
  detection. In: The IEEE International Conference on Computer Vision (ICCV)
  (October 2019)

\end{thebibliography}
\end{document}